\documentclass[journal]{IEEEtran}
\usepackage{graphicx}

\input{tcilatex}

\begin{document}

\title{End-to-End Deep Learning of Lane Detection and Path Prediction for Real-Time Autonomous Driving}

\author{Der-Hau~Lee and~Jinn-Liang~Liu \thanks{%
This work was supported by the Ministry of Science and Technology, Taiwan,
under Grant MOST 109-2115-M-007-011-MY2.} \thanks{%
D. Lee was with the Department of Electrophysics, National Chiao Tung
University, Hsinchu 300, Taiwan.} \thanks{%
J. Liu is with the Institute of Computational and Modeling Science, National
Tsing Hua University, Hsinchu 300, Taiwan (e-mail:
jinnliu@mail.nd.nthu.edu.tw, website: http://www.nhcue.edu.tw/~jinnliu).} }
\maketitle

\begin{abstract}
Inspired by the UNet architecture of semantic image segmentation, we propose a lightweight
UNet using depthwise separable convolutions (DSUNet) for end-to-end learning of lane
detection and path prediction (PP) in autonomous driving. We also design and integrate a
PP algorithm with convolutional neural network (CNN) to form a simulation model (CNN-PP)
that can be used to assess CNN's performance qualitatively, quantitatively, and dynamically
in a host agent car driving along with other agents all in a real-time autonomous manner.
DSUNet is 5.16x lighter in model size and 1.61x faster in inference than UNet. DSUNet-PP
outperforms UNet-PP in mean average errors of predicted curvature and lateral offset for
path planning in dynamic simulation. DSUNet-PP outperforms a modified UNet in lateral error,
which is tested in a real car on real road. These results show that DSUNet is efficient
and effective for lane detection and path prediction in autonomous driving.
\end{abstract}


\section{Introduction}\label{sec1}

Lane detection is an essential task of advanced driver assistance systems
(ADAS) in modern vehicles and autonomous driving systems in self-driving
cars for lane keeping assistance, departure warning, and centering \cite{Huv15,Bru18,Gar19,Kim18,Lin00,Cud20,Bad21,Li20,Zou20,Lu21}. There are two major paradigms,
namely, classic computer vision and deep learning, for lane detection \cite{Zou20}. The classic computer vision paradigm requires intensive feature
engineering, road modeling, and special case handling and is thus not robust
enough to deal with seemingly infinite driving situations, environments, and
unexpected obstacles \cite{Huv15}. The deep learning paradigm has
shown considerable progress to alleviate these difficulties in recent years \cite{Zou20}.

Path prediction (lane following) is also a crucial task of ADAS and
self-driving vehicles for active driving assistance and lateral and
longitudinal controls \cite{Kim18,Lin00,Cud20,Bad21}. It depends on various factors of
vehicles and road and traffic environment such as ego vehicle's speed and
steering angle, road geometry, and the other closest in-path vehicle \cite{Kim18,Lin00,Cud20,Bad21}. It also involves perception, sensory, and control systems in
the ego vehicle \cite{Lin00,Kim18}.

UNet \cite{Ron15} is an encoder-decoder convolutional neural network (CNN)
\cite{Hin06} for semantic segmentation. It achieved remarkable success in many
competitions \cite{Liu20} due to its elegant architecture of symmetrical contracting
(encoding) and expanding (decoding) sub-networks with skip connections that combine
semantic (global) and appearance (local) information from both sub-networks \cite{Lon15}.

The accuracy and efficiency of CNNs are equally important in a variety of applications such as robotics, mobile devices, and self-driving cars \cite{Ian16,Cho17,How17,Zha17,San18,Hus19,Beh20,Gad20,Gri20,Lee21}. The concept of depthwise separable convolutions \cite{Sif14} (a depthwise convolution followed by a pointwise convolution \cite{Cho17}) has been used to construct many efficient CNNs  \cite{Ian16,Cho17,How17,Zha17,San18,Beh20,Gad20}. We apply this concept to UNet \cite{Ron15} (DSUNet) for semantic road segmentation and lane detection and integrate DSUNet (or other CNNs) with path prediction (PP) algorithm to form a simulation model (CNN-PP) that can be used to assess CNN's performance qualitatively, quantitatively, and dynamically in a host agent car driving along with other agents all in a real-time autonomous manner \cite{Lee21,Che15}.

The small segmantation network DSUNet not only has real-time performance with merits on energy and memory consumption but also attains sufficient accuracy in dynamic traffic. We show that DSUNet is 5.16x lighter and 1.61x faster than UNet with comparable accuracy.

\begin{figure}
\centering
\includegraphics[width=0.45\textwidth]{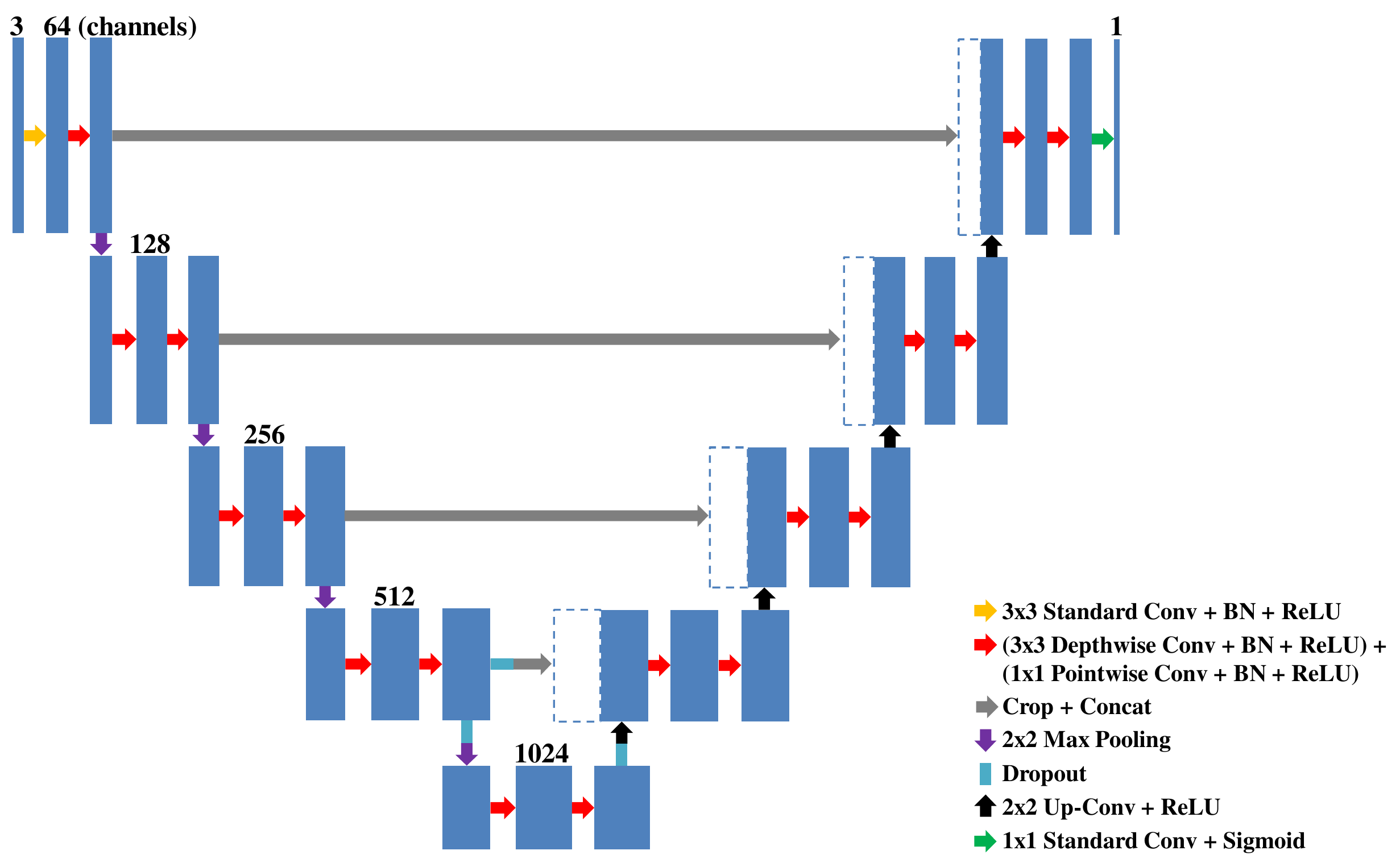}
\caption{DSUNet replaces 17 3x3 standard conv layers in UNet \cite{Ron15} with 17 3x3 depthwise and 17 1x1 pointwise conv layers shown by red arrows in the architecture with 3 dropouts by light-blue bars. Other operations remain the same. The ratios of DSUNet/UNet are 40/23 conv layers, 6.01/31.04 M parameters, 9.56/62.51 B MACs, and 45/28 FPS in inference.}
\end{figure}

\section{Related work}

Fig. 1 illustrates the architecture of both the original UNet \cite{Ron15} and DSUNet. DSUNet replaces 17 3x3 convolutional (conv) layers in UNet with 17 3x3 depthwise and 17 1x1 pointwise conv layers shown by red arrows in the figure with 3 dropouts by light-blue bars. The batch normalization and padding method of \cite{Bru18} are used in DSUNet. The input and output channels of DSUNet/UNet are 3/1 and 1/2, respectively, due to different input images (color/black and white) and output
loss functions (weighted cross entropy/softmax). Other operations remain the same. The ratios of DSUNet/UNet are 40/23 conv layers, 6.01/31.04 M parameters, 9.56/62.51 B multiplication accumulation operations (MACs) with the same I/O of DSUnet, and 45/28 frames per second (FPS) in inference on a PC with Intel i9-9900K and an NVIDIA RTX 2080ti.

In the past few years, UNet have been modified or extended for its use in various applications \cite{Liu20} such as 3D volumetric segmentation \cite{Cic16}, 3D stroke lesion segmentation \cite{Tur18}, retinal vessel segmentation \cite{Xia18,Li19}, and photoacoustic imaging \cite{Gua20}. Bruls et al. first applied UNet to road marking detection with batch normalization after every convolutional layer \cite{Bru18}. Zou et al. replaced the bottom layer of UNet with a convolutional long short-term
memory network to form UNet-ConvLSTM, where UNet features of continuous multiple frames are inputs to LSTM in time series for feature learning and lane prediction \cite{Zou20}. UNet-ConvLSTM improves inference accuracy significantly, especially for road images with worn-up or occluded lane markings.

In addition to CNN's performance in accuracy, its computational complexity is a critical concern in self-driving cars \cite{Hus19,Gri20}. Depthwise separable convolutions \cite{Sif14} separate the operations of spatial and channel correlations \cite{Cho17} using single-channel kxk and 1x1 filters \cite{Ian16}, respectively, and thus greatly reduce parameters \cite{Ian16,Cho17}. This idea inspired a variety of efficient models such as SqueezeNet, Xception, MobileNet, and ShuffleNet derived from AlexNet, GooLeNet, or ResNet for classification or object detection \cite{Ian16,Cho17,How17,Zha17}.

From UNet for semantic segmentation, Beheshti and Johnsson designed Squeeze-UNet \cite{Beh20} by replacing the contracting and expanding layers with fire modules inspired by SqueezeNet \cite{Ian16}. The ratios of Squeeze-UNet/UNet are  0.8 to 0.98 in accuracy for five classes (car, road, tree, building, and sky), 1/12 in model size MB, 1/3.2 in MACs, and 0.83 and 0.48 in inference and training time, respectively. The main differences between our and their model are the ratios (DSUnet/Squeeze-UNet) 0.46 and 1.33 in MACs and the speed of inference, respectively, i.e., DSUnet is 1.33x faster. We use standard depthwise separable convolutions \cite{Ian16,Cho17,How17} instead of fire modules that repeatedly squeeze (using 1x1 filters) and expand (mixing 1x1 and 3x3 filters) conv layers.

Gadosey et al. also applied DS convolutions with weight standardization and group normalization to UNet (SD-UNet) for semantic segmentation of biomedical images, where SD is short for stripping down \cite{Gad20}. Their results show that the ratios of SD-UNet/UNet are 0.97/0.98  for accuracy, 3.9/31 M parameters, 7.8/62.04 floating point operations, and 91/56 in inference speed. The ratios of DSUNet/SD-UNet are 1 for accuracy, 1.53 in parameters, 1.22 in operations, and 0.99 in speed. SD-UNet is much lighter than DSUNet in computational complexity. Nevertheless, the performance of these two models in accuracy and speed is almost the same on the baseline of UNet for two different applications.

\begin{figure}
\centering
\includegraphics[width=0.45\textwidth]{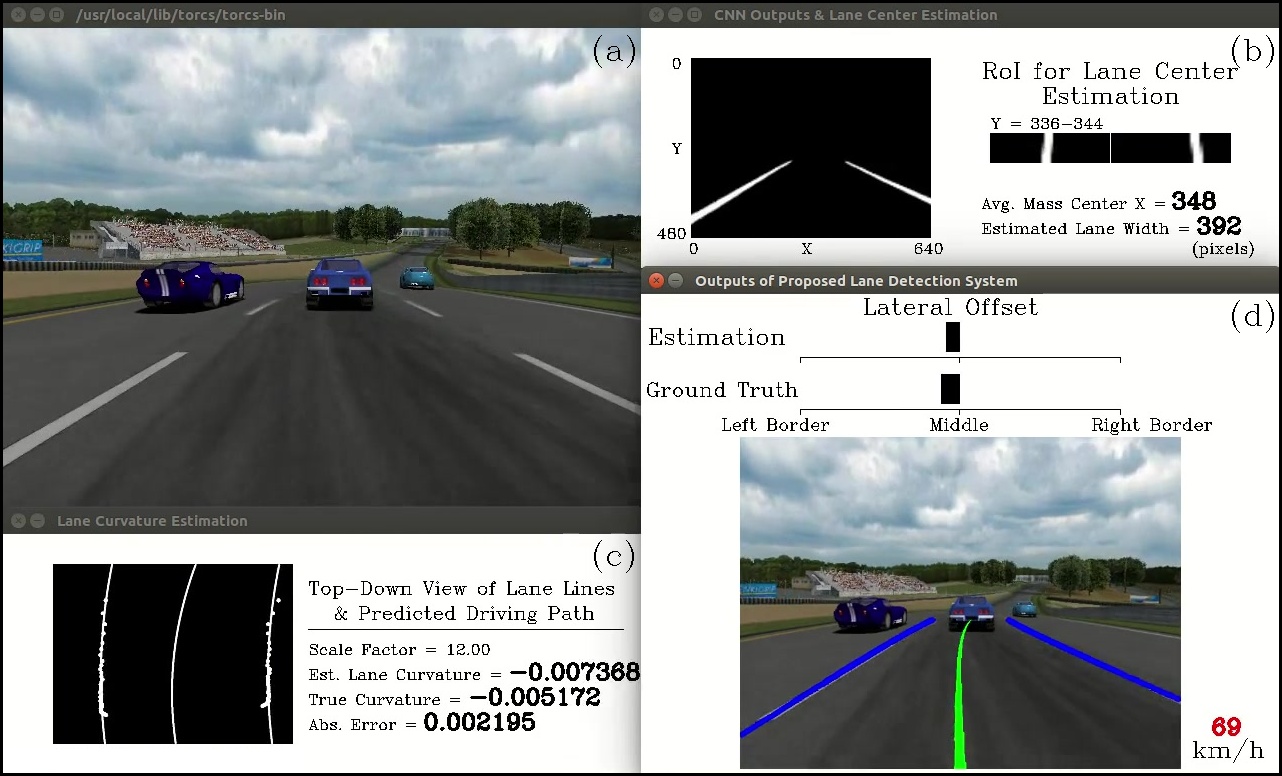}
\caption{Real-time simulation
architecture of lane detection and path prediction by DSUNet-PP from a host
car's view in TORCS traffic: (a) input image, (b) lane detection, (c) line clustering and perspective transformation, and (d) lateral offset
estimation, inverse perspective transformation, line (blue) fitting, and
driving path (green) prediction}
\end{figure}

Path prediction models are generally polynomials of lane centerline having
few coefficients determined by heading and lane curvature without \cite{Cud20} or with considering vehicle's transient dynamics such as yaw rate
and lateral and longitudinal velocities \cite{Lin00}. Cudrano et al. \cite{Cud20} proposed an integrated NN-PP model that uses vehicle's heading and
lateral displacement calculated from the lane detection output of a modified
UNet to determine a cubic polynomial. They tested the model in a real
car driving on two racetracks (one is more curvy and the other is less)
without other cars and with a maximum speed of 54 km/h. They tested PP algorithm with results but did not present the overall
architecture of NN-PP model, the details of the modified UNet, and its performance on lane detection.

We proposed a deep learning model in \cite{Lee21} that combines multi-task
CNN and control algorithms for multiple actions in autonomous driving
(steering, acceleration, braking, lane changing, and overtaking). We also
modified The Open Racing Car Simulator (TORCS) \cite{Che15,Lee21,Wym00} on which the model was implemented for a
host car (Host) efficiently inferring and stably driving with other
autonomous cars (Agents) in dynamic TORCS traffic \cite{Lee21,Che15}. We extend here our TORCS
to include lane detection and path prediction in order to test and evaluate
the overall performance of deep learning, lane detection, and path
prediction algorithms in the simulator before deploying them to real cars in
real world traffic.

\subsection{Contributions}

Scarce literature exists on lightweight CNN and path
prediction algorithms for self-driving cars to perform lane detection, road
recognition, and lane following in dynamic traffic \cite{Cud20}, let alone
the overall performance of these algorithms in a real-time autonomous manner. The contributions of the present work are summarized as follows:

(i) We propose DSUNet, a lightweight CNN, that extends the original UNet from classification and object detection to semantic segmentation and improves UNet by 5.12x, 6.54x, and 1.61x in model size, MACs, and FPS, respectively.

(ii) We propose an end-to-end CNN-PP model for simulating self-driving cars and evaluating the dynamical performance of CNN and PP algorithms in real time.

(iii) We show that DSUNet-PP outperforms UNet-PP in both static and dynamic mean average errors of predicted curvature and lateral offset (vehicle center to lane center \cite{Jun05}) for path planning in dynamic traffic of self-driving agents with lane changes and a max speed of 70 km/h.

(iv) We show that DSUNet-PP outperforms the modified UNet of \cite{Cud20} in lateral error on different (real-world \cite{Cud20} vs. curvier artificial) roads by different
(real \cite{Cud20} vs. artificial) cars.

\section{Deep learning of lane detection and path prediction}

Fig. 2 illustrates the architecture of our deep learning, lane detection, and path prediction
algorithms implemented in TORCS from Host's view with several Agents in
dynamic traffic. The architecture consists of (a) images from Host's front-facing camera, (b) gray lane images obtained from a CNN (DSUNet for example in Fig. 2) and their
region of interest (RoI) for lane localization and centering, (c) a
perspective mapping (PM) of lane lines and their clustering from camera's
view (CV) to top-down view (TDV) for path prediction and curvature
estimation, and (d) the inverse PM lane lines (blue) and prediction path (green) with lateral offset estimation from TDV back to CV. More details about these four
sub-figures and corresponding methods are given below.

To handle the unbalanced
numbers of positive (lane line) and background samples, we use the weighted cross entropy loss \cite{Wan17}
\begin{eqnarray}
L = &-& \frac{N}{P+N}\sum\limits_{y_c=1} \log \left(\sigma(x_c) \right)\notag \\
&-& \frac{P}{P+N} \sum\limits_{y_c=0} \log \left(1-\sigma(x_c) \right)
\end{eqnarray}
for training CNNs, where ${x_c }$ is a sample score, ${y_c }$ is the corresponding label, $\sigma$ is the sigmoid function, and $P$ ($N$) is the number of positive (negative) samples in a batch of images.

\subsection{Lateral offset estimation}

Cameras mounted on vehicles vary in height and lateral location so do lane
images \cite{Lu21}. We first resize the output gray (probability) images to the
original size 640x480 and then choose the region of interest
(RoI shown in Fig. 2b) of gray images in the range of $y= 336 \sim 344$ for lane centering by calculating the centroid coordinate $\overline{x}$. The coordinates
of left ($x = 0 \sim 320$) and right ($x = 321 \sim 639$) parts of
gray images are $\overline{x}_{l}$ and $\overline{x}_{r}$, respectively. The lane center is estimated as $\left(\overline{x}_{l}+\overline{x}_{r}\right) /2$ and the lane width is $\overline{x}_{r}-\overline{x}_{l}$. Since the image center is at $x=320$, the distance from Host to the lane center is estimated by $\Delta _{0}=320-\left(\overline{x}_{l}+\overline{x}_{r}\right) /2$.

However, this approach tends
to overestimate the true distance to the lane center since RoI is ahead of
Host and depends on camera's position. We use $\Delta =\alpha \Delta _{0}$
with $\alpha =0.6$ as an estimated offset to the true lane center shown in
Fig. 2d. This lane centering is thus a CNN-based method.

\subsection{ Driving path prediction}

The following path prediction algorithm comprises well-known or advanced
post-processing methods that are particularly suitable for output images
having local and global properties to deal with complexity,
scalability, and homography issues \cite{Li20,Lu21}.

\textit{Path Prediction Algorithm:}

\textit{Step 1.} We use DBSCAN
\cite{Huv15} to cluster detected line segments (dots in Fig. 2c).

\textit{Step 2.} A perspective mapping \cite{Li20,Lu21} then
transforms semantic lanes from camera's view to top-down view space to which
quadratic polynomials \cite{Gar19} apply for fitting parallel lane lines
(Fig. 2c) by using recursive least squares \cite{Cud20} and imposing
parallelism.

\textit{Step 3.} Another quadratic polynomial $f(x)$ for the desired driving
path is uniquely determined by three conditions, i.e., the starting ($x_{s}$%
) and final ($x_{f}$) points of the middle line (Fig. 2c) of two parallel
lines and its tangent at $x_{s}$ as that of the parallel lines. We thus
obtain the TDV lane curvature $\kappa _{0}=f^{\prime \prime }(x_{s})/\left[
1+f^{\prime }(x_{s})\right] ^{3/2}$ at $x_{s}$ in 1/m unit. The true
curvature $\kappa =\beta \kappa _{0}$ in the real world space with a scale
factor $\beta =12$ is then used to calculate its error compared to the
ground truth curvature of TORCS track as shown in Fig. 2c.

\textit{Step 4.} Finally, the inverse PM transforms these three lines back
to the CV space for visual verification and lateral estimation shown in Fig.
2d.

In Step 1, we use five consecutive gray images to obtain averaged line
segments for possible occluded or worn-out markings. Higher-order
polynomials and corresponding PM transformations are generally needed in
Steps 2 and 3 for curvy or up-and-down roads in real world \cite{Lu21}.

\section{Experimental setup}

Lane segmentation results predicted by a CNN are expressed in pixel-wise true positive, true negative, false positive, and
false negative values which are paired with ground truth values of lane lines in images. Its performance is measured in accuracy, precision, recall,
and F1 score \cite{Zou20}.

The image data of this work consist of artificial (TORCS) and real-world (LLAMAS \cite{Beh19} and TuSimple \cite{Tus21}) images summarized in Table 1, where TuSimple includes images augmented by mirroring in training phase. The ratio between the numbers of training and test images is about 10. We annotate TORCS images (collected by Host driving on Tracks 1 to 6 \cite{Che15}) to binary images of ego-lane lines, which are similar to those annotated from real-world images \cite{Beh19}. A second test set of 3,200 images from curvy Track 7 \cite{Che15} is used to evaluate the overall performance (Figs. 2b-2d) of the integrated CNN-PP model from Host's perspective in real-time and dynamic traffic with other Agents.

\begin{table}[!t]
\caption{Image data of this work}
\begin{center}
\begin{tabular}{c|c|c|c}
\hline
Dataset & TORCS & LLAMAS  & TuSimple   \\ \hline
No. of images & 42747 & 22714 & 6408   \\\hline
Labels  & \multicolumn{2}{c|}{ego-lane lines}      &   multi-lane lines   \\ \hline
Source & this work & \cite{Beh19} & \cite{Tus21}   \\\hline
\end{tabular}
\end{center}
\end{table}

UNet and DSUNet are trained on separate TORCS, LLAMAS, TuSimple, and TuSimple-mirror images with Adam optimizer, a batch size of 1, momentum 0.9, and learning rate $10^{-4}$ and $10^{-5}$ for first 75 and last 25 epochs, respectively. The trained models are evaluated on the first test set for their performance of lane detection. We then evaluate the static and dynamic performance (as shown in Fig. 2) of the integrated CNN-PP model in TORCS for Host driving with 6 Agents on Track 7 having a total length of 3919 m. All cars drive autonomously by default control algorithms \cite{Lee21} with the maximum speed of 70 and 67 km/h for Host and Agents, respectively, as displayed in
Fig. 2d.

\begin{table}[!t]
\caption{Performance of UNet and DSUNet for lane detection on testing data}
\begin{center}
\begin{tabular}{c|c|c|c|c}
\hline
\multicolumn{5}{c}{TORCS}  \\ \hline
 & Accuracy & Precision & Recall & F1 score  \\ \hline
UNet    & 0.995	& 0.933	& 0.911	& 0.921 \\ \hline
DSUNet & 0.990	& 0.902	& 0.833	& 0.858 \\ \hline
\multicolumn{5}{c}{LLAMAS}  \\ \hline
 & Accuracy & Precision & Recall & F1 score  \\ \hline
UNet    & 0.986	& 0.789	& 0.910	& 0.843 \\ \hline
DSUNet & 0.984	& 0.768	& 0.902	& 0.825\\ \hline
\multicolumn{5}{c}{TuSimple}  \\ \hline
 & Accuracy & Precision & Recall & F1 score  \\ \hline
UNet    & 0.974	& 0.873	& 0.758	&  0.809\\ \hline
DSUNet & 0.971& 0.817& 0.786& 0.796 \\ \hline
\multicolumn{5}{c}{TuSimple-mirror }  \\ \hline
 & Accuracy & Precision & Recall & F1 score  \\ \hline
UNet    & 0.974   & 0.821 & 0.826 & 0.822  \\ \hline
DSUNet & 0.967& 0.750& 0.841& 0.789\\ \hline
\end{tabular}
\end{center}
\end{table}

\section{Results}

Table 2 shows that DSUNet is comparable to UNet in four measures for lane detection on almost all images (up to 96\% in average F1) in the first test dataset as illustrated in Fig. 3, for example.

\begin{figure}
\centering
\includegraphics[width=0.45\textwidth]{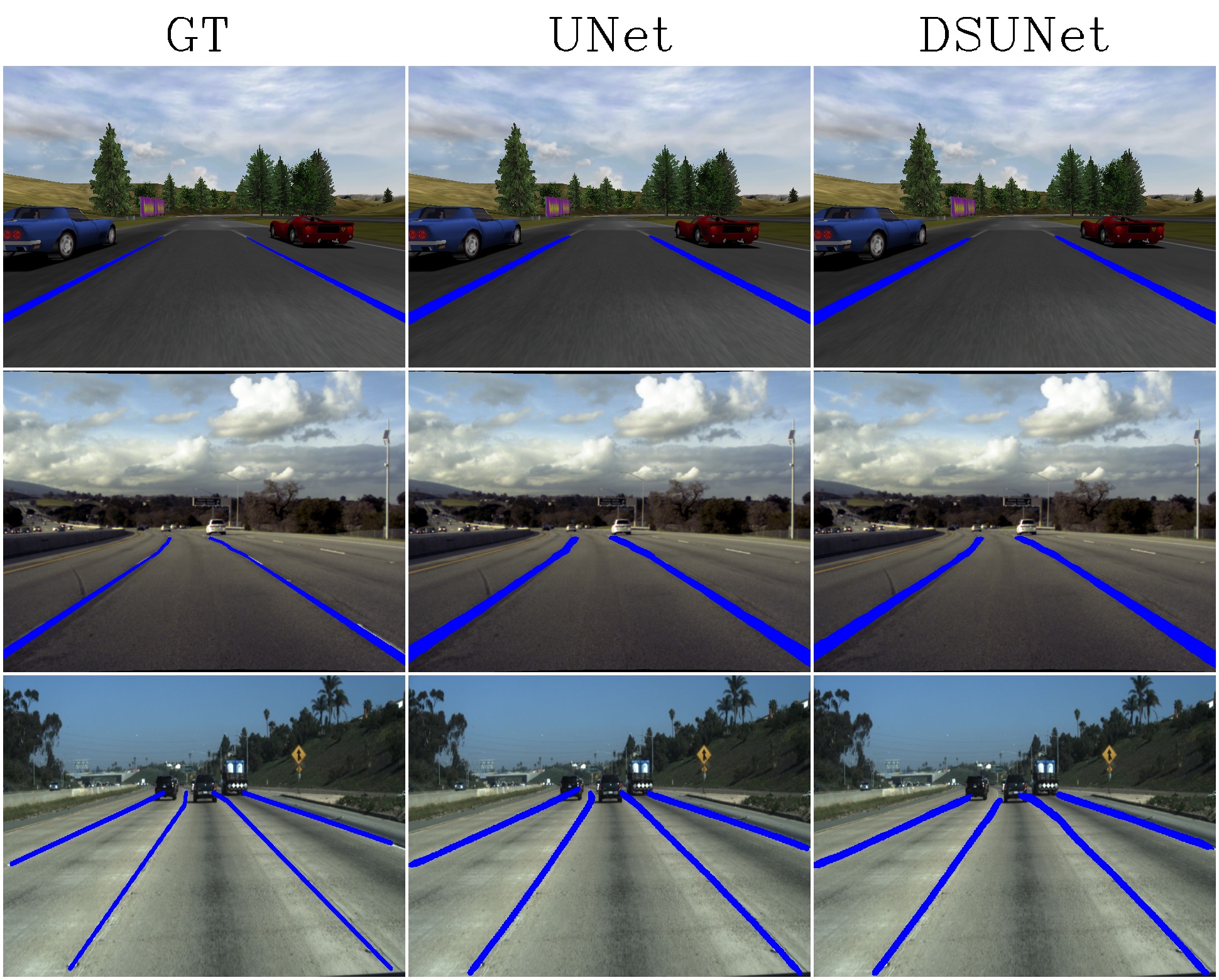}
\caption{Comparison of UNet (second column) and DSUNet (third column) in lane detection with ground truth (first column) on TORCS (first row), LLAMAS \cite{Beh19} (second row), and TuSimple \cite{Tus21} (third row) images}\label{fig4}
\end{figure}

\begin{table}
\caption{Performance of UNet-PP and DSUNet-PP in static and dynamic errors (sMAE/dMAE) with the percentage (\%) of available frames (Avail) on Track 7 \cite{Che15}}
\begin{center}
\begin{tabular}{c|c|c}
\hline
& UNet-PP & DSUNet-PP \\ \cline{2-3}
\multicolumn{1}{l|}{} & \multicolumn{2}{|c}{sMAE/dMAE} \\ \hline
\multicolumn{1}{l|}{$\kappa $ (1/m)} & 0.0048/0.0052 & 0.0046/0.0049 \\
\multicolumn{1}{l|}{$\Delta $ (m)} & 0.1909/0.1125 & 0.1771/0.1018 \\ \hline
\multicolumn{1}{l|}{} & \multicolumn{2}{|c}{Avail (\%)} \\ \hline
\multicolumn{1}{l|}{$\kappa $} & 100/100 & 100/100 \\
\multicolumn{1}{l|}{$\Delta $} & 95.68/94.06 & 83.50/82.06 \\ \hline
\end{tabular}
\end{center}
\end{table}

Table 3 shows the overall performance of UNet-PP and DSUNet-PP measured in the static and real-time dynamic mean absolute errors
(sMAE/dMAE) \cite{Lee21} of predicted curvature $\kappa$ (in 1/m) and
lane center distance $\Delta$ (m) on the path traversed by Host in TORCS
traffic with other Agents. Here, dMAE means that MAE is dynamically calculated while Host is in motion as shown in Fig. 2c in contrast to sMAE from recorded images. The percentage (\%) of available frames (Avail) \cite{Cud20} from Host's videos for predicting $\Delta$ is less than 100 due to shadows or occlusions on the road (RoI in Fig. 2b is not available) and is worse in dynamic than in static conditions.

\begin{figure}
\centering
\includegraphics[width=0.45\textwidth]{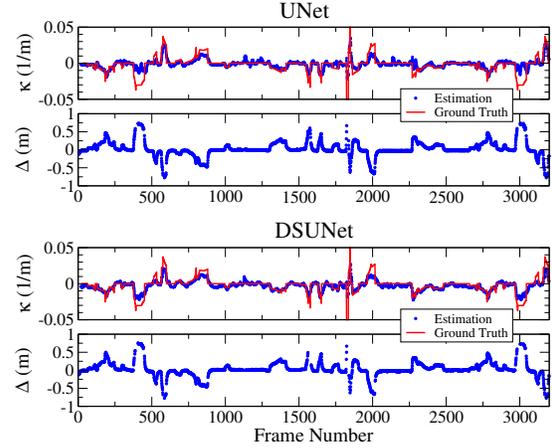}
\caption{Estimated curvatures ($\protect\kappa$) and distances to lane center ($\Delta $) in blue by UNet-PP and DSUNet-PP with ground-truth in red on a sequence of frames from Track 7}
\end{figure}

DSUNet is surprisingly better than UNet in both sMAE and dMAE for both $\kappa$ and $\Delta$ although it is much smaller in model size. Its available frames for calculating $\Delta$ dMAE have been reduced by 19.03\%. We attribute its better performance to more missing frames for calculating $\Delta$ as these obscure frames would increase its $\Delta$ MAE if they were accounted for. $\Delta$ dMAE is smaller than $\Delta$ sMAE for the same reason. $\Delta$ and $\kappa$ MAEs are correlated in CNN-PP model.

Table 3 thus shows
important effects of CNN's perception in motion on path planning in autonomous driving. Our model performs better than that of \cite{Cud20} (cf. Table 3), i.e., 10.18 vs. 45.3 cm in $\Delta$ dMAE in centered driving style on different (artificial vs. real-world) roads by different
(artificial vs. real) cars at different speeds (70 vs. 54 km/h). Cudrano et al. \cite{Cud20} also modified UNet
and integrated it in a perception pipeline that includes $\Delta $
calculations but without using learning methods for lateral corrections like
our lateral offset estimation.

Fig. 4 shows the ground-truth (red) and estimated (blue dotted)
curves of $\kappa$ and $\Delta$ in sMAE with respect to the frame number
of a video from Track 7 recorded by Host driving alone in the middle
lane, where each point of the curves is an average of 11 values of $\kappa$
or $\Delta$ from consecutive frames. The $\kappa$ and $\Delta$ curves are
reciprocal in peaks and valleys as in their units. The curvature ($\kappa$) at current frame (time) is estimated by the one-dimensional Kalman filter \cite{Wel01} using consecutive 15 noisy data points as curvature measurements, which are numerical values from Step 3 in the PP algorithm, where the covariance of measurement and process noises is 0.01 and 0.0001, respectively. DSUNet-PP outperforms UNet-PP in deviation from ground-truth values on several curvy sections (near frame numbers
400, 2200 and 3000, for example).

The simulation track is more curvy than real or experimental roads \cite{Lu21} by one order of magnitude. The minimum radius of general roads is about 130-160 m \cite{Fit94} corresponding to a maximum curvature of 0.00625-0.0076 ${\rm m}^{ - 1}$. Table 3 and Fig. 4 thus postulate that DSUNet-PP may reduce the dynamic error of its lateral ($\Delta$) prediction to 5 cm on real roads in future studies. Two simulation videos are available on our website for a qualitative and quantitative review on the dynamic performance of UNet-PP and DSUNet-PP from Host driving
on Track 7 with other Agents and lane changes in real time.

\section{Conclusion}
Based on UNet and depthwise separable convolutions, we proposed a lightweight DSUNet for lane detection, road recognition, and path prediction (PP). We also proposed a model that integrates CNN and PP algorithms and developed a simulator to evaluate the dynamic performance
of the model qualitatively as well as quantitatively from a host car driving
autonomously with other cars and lane changes. Our results show that
(i) DSUNet improves UNet by 5.12x, 6.54x, and 1.61x in model size, MACs, and FPS, respectively, (ii) DSUNet-PP outperforms UNet-PP in both static and dynamic mean average errors of predicted curvature and lateral offset for path planning in dynamic traffic, and (iii) DSUNet-PP outperforms the modified UNet of \cite{Cud20} in lateral error on different (artificial vs. real-world \cite{Cud20}) roads by different (artificial vs. real) cars.



\begin{thebibliography}{1}

\bibitem{Huv15} Huval, B., et al.: An empirical evaluation
of deep learning on highway driving. arXiv preprint {arXiv}:1504.01716 (2015)

\bibitem{Bru18} Bruls, T., Maddern, W., Morye, A.A., Newman, P.: Mark yourself: Road marking segmentation via weakly-supervised annotations from multimodal data. {IEEE International Conference on Robotics and Automation (ICRA)}, pp. 1863-1870 (2018)

\bibitem{Gar19} Garnett, N., et al.: 3D-LaneNet: End-to-end 3D multiple lane detection. In: {Proceedings of the IEEE International Conference on Computer Vision}, pp. 2921-2930 (2019)

\bibitem{Kim18} Kim, W., et al.: Vehicle path prediction using yaw acceleration for adaptive cruise control. {IEEE Transactions on Intelligent Transportation Systems}. \textbf{19}, 3818-3829 (2018)

\bibitem{Lin00} Lin, C.F., Ulsoy, A.G., LeBlanc, D.J.: Vehicle dynamics and external disturbance estimation for vehicle path prediction. {IEEE Transactions on Control Systems Technology}. \textbf{8}, 508-518 (2000)

\bibitem{Cud20} Cudrano, P., et al.: Advances in centerline estimation for autonomous lateral control. In: {IEEE Intelligent Vehicles Symposium (IV)}, pp. 1415-1422 (2020)

\bibitem{Bad21} Badue, C., et al.: Self-driving cars: A survey. Expert Systems with Applications. \textbf{165}, 113816 (2021)

\bibitem{Li20} Li, X., Li, J., Hu, X., Yang, J.: Line-CNN: End-to-end traffic line detection with line proposal unit. {IEEE Transactions on Intelligent Transportation Systems}. \textbf{21}, 248-258 (2020)

\bibitem{Zou20} Zou, Q., et al.: Robust lane detection from continuous driving scenes using deep neural networks. {IEEE Transactions on Vehicular Technology}. \textbf{69}, 41-54 (2020)

\bibitem{Lu21} Lu, P., et al.: SUPER: A novel lane detection system. {IEEE Transactions on Intelligent Vehicles}. \textbf{6}, 583-593 (2021)

\bibitem{Ron15} Ronneberger, O., Fischer, P., Brox, T.: U-Net: Convolutional networks for biomedical image segmentation. In: {International Conference on Medical Image Computing and Computer-Assisted Intervention (MICCAI)}, pp. 234-241 (2015)

\bibitem{Hin06} Hinton, G.E., Salakhutdinov, R.R.: Reducing the dimensionality of data with neural networks. {Science}. \textbf{313}, 504-507 (2006)

\bibitem{Liu20} Liu, L., et al.: A survey on U-shaped networks in medical image segmentations. {Neurocomputing}. \textbf{409}, 244-258 (2020)

\bibitem{Lon15} Long, J., Shelhamer, E., Darrell, T.: Fully convolutional networks for semantic segmentation. In: {Proceedings of the IEEE conference on computer vision and pattern recognition}, pp. 3431-3440 (2015)

\bibitem{Ian16} Iandola, F.N., et al.: SqueezeNet: AlexNet-level accuracy with 50x fewer parameters and $<$0.5MB model size. arXiv preprint {arXiv}:1602.07360 (2016)

\bibitem{Cho17} Chollet, F.: Xception: Deep learning with
depthwise separable convolutions. In: {Proceedings of the IEEE conference on computer vision and pattern recognition}, pp. 1251–1258 (2017)

\bibitem{How17} Howard, A.G., et al.: MobileNets: Efficient convolutional neural networks for mobile vision applications. arXiv preprint {arXiv}:1704.04861 (2017)

\bibitem{Zha17} Zhang, X., Lin, M., Sun, J.: ShuffleNet: An extremely efficient convolutional neural network for mobile devices. arXiv preprint {arXiv}:1707.01083 (2017)

\bibitem{San18} Santos, A.G., et al.: Reducing SqueezeNet storage size with depthwise separable convolutions. In: {International Joint Conference on Neual Networks (IJCNN)}, pp. 1-6 (2018)

\bibitem{Hus19} Hussain, R., Zeadally, S.: Autonomous cars: Research results, issues, and future challenges. {IEEE Commun. Surv. Tutor.} \textbf{21}, 1275-1313 (2019)

\bibitem{Beh20} Beheshti, N., Johnsson, L.: Squeeze U-Net: A memory and energy efficient image segmentation network. In: {IEEE/CVF Conference on Computer Vision and Pattern Recognition Workshops (CVPRW)}, pp. 1495-1504 (2020)

\bibitem{Gad20} Gadosey, P.K., et al.: SD-Unet: Stripping down U-net for segmentation of biomedical images on platforms with low computational budgets. {Diagnostics}. \textbf{10}, 110 (2020)

\bibitem{Gri20} Grigorescu, S., Trasnea, B., Cocias, T., Macesanu, G.: A survey of deep learning techniques for autonomous driving. {J. Field Robotics.} \textbf{37}, 362-386 (2020)

\bibitem{Lee21} Lee, D.-H., et al.: Deep learning and control algorithms of direct perception for autonomous driving. {Applied Intelligence.} \textbf{51}, 237-247 (2021)

\bibitem{Sif14} Sifre, L.: Rigid-motion scattering for image classification. {Ph. D. thesis}, Ecole Polytechnique, Palaiseau, France (2014)

\bibitem{Che15} Chen, C., Seff, A., Kornhauser, A., Xiao, J.: DeepDriving: Learning affordance for direct perception in autonomous driving. In: {IEEE International Conference on Computer Vision (ICCV)}, pp. 2722-2730 (2015)

\bibitem{Cic16} {\c C}i{\c c}ek, {\"O}., et al.: 3D U-Net: Learning dense volumetric segmentation from sparse annotation. In: {MICCAI 2016: Medical Image Computing and Computer-Assisted Intervention}, pp. 424-432 (2016)

\bibitem{Tur18} Rogriguez-Sanchez, A., Ture\v{c}kov{\'a}, A.: Isles challenge: U-shaped convolution neural network with dilated convolution for 3d stroke lesion segmentation. In: {International MICCAI Brainlesion Workshop}, pp. 319–327 (2018)

\bibitem{Xia18} Xiao, X., Lian, S., Luo, Z., Li, S.: Weighted Res-UNet for high-quality retina vessel segmentation. In: {9th International Conference on Information Technology in Medicine and Education (ITME)}, pp. 327-331 (2018)

\bibitem{Li19} Li, R., Li, M., Li, J., Zhou, Y.: Connection sensitive attention U-NET for accurate retinal vessel segmentation. arXiv preprint {arXiv}:1903.05558 (2019)

\bibitem{Gua20} Guan, S., Khan, A.A., Sikdar, S., Chitnis, P.V.: Fully dense UNet for 2-D sparse photoacoustic tomography artifact removal. {IEEE Journal of Biomedical and Health Informatics.} \textbf{24}, 568-576 (2020)

\bibitem{Wym00} Wymann, B., et al.: TORCS: The open racing car simulator. (2000)

\bibitem{Jun05} Jung, C.R., Kelber, C.R.: A lane departure warning system using lateral offset with uncalibrated camera. In: Proceedings IEEE  Intelligent Transportation Systems, pp. 102-107 (2005)

\bibitem{Wan17} Wang, X., et al.: ChestX-ray8: Hospital-scale chest X-ray database and benchmarks on weakly-supervised classification and localization of common thorax diseases. In: {IEEE Conference on Computer Vision and Pattern Recognition(CVPR)}, pp. 3462-3471 (2017)

\bibitem{Beh19} Behrendt, K., Soussan, R.: Unsupervised Labeled Lane Markers Using Maps. In: {Proceedings of the IEEE/CVF International Conference on Computer Vision Workshops (ICCVW)}, pp. 832-839 (2019)

\bibitem{Tus21} TuSimple Competitions for CVPR2017. https:// github.com/TuSimple /tusimple-benchmark. Accessed 11 Aug 2021

\bibitem{Fit94} Fitzpatrick, K.: Horizontal Curve  Design: An Exercise in Comfort and Appearance. {Transportation Research Record.} \textbf{1445}, 47-53 (1994)

\bibitem{Wel01} Welch, G., Bishop, G.: An introduction to the Kalman filter. Univ. North Carolina, Chapel Hill, NC, USA, Lecture (2001)

\end{thebibliography}
\end{document}